\documentclass{article}

\usepackage[final]{colm2026_conference}

\usepackage{microtype}
\usepackage{graphicx}
\usepackage{booktabs}
\usepackage{amsmath}
\usepackage{amssymb}
\usepackage{hyperref}
\usepackage{url}
\usepackage{xspace}
\usepackage{multirow}
\usepackage{array}
\usepackage{longtable}
\usepackage{lineno}

\definecolor{darkblue}{rgb}{0, 0, 0.5}
\hypersetup{
  colorlinks=true,
  citecolor=darkblue,
  linkcolor=darkblue,
  urlcolor=darkblue,
  pdftitle={Reach Into The CHOIR: Free-List Elicitation Uncovers Distinct Model Voices in LLM Ensembles},
  pdfauthor={Ben Wigler and Maria Tsfasman}
}

\newcommand{\choir}{\textsc{CHOIR}\xspace}
\newcolumntype{P}[1]{>{\raggedright\arraybackslash}p{#1}}

\title{Reach Into The CHOIR: Free-List Elicitation Uncovers\\Distinct Model Voices in LLM Ensembles}

\author{%
Ben Wigler \quad Maria Tsfasman\\
\textit{LoveMind AI}, New York, USA\\
\href{mailto:benben@lovemind.ai}{benben@lovemind.ai} \quad
\href{mailto:masha@lovemind.ai}{masha@lovemind.ai}\\
\href{https://orcid.org/0009-0009-4406-187X}{ORCID: 0009-0009-4406-187X} \quad
\href{https://orcid.org/0000-0001-5582-7636}{ORCID: 0000-0001-5582-7636}
}

\begin{document}

\ifcolmsubmission
\linenumbers
\fi

\maketitle
\raggedbottom

\begin{abstract}
Open-ended LLM homogeneity can create false plurality when several systems appear to offer independent perspectives while returning the same familiar default. Single-pass answers obscure the distinction between agreement produced by a tightly constrained answer space, prompt-vocabulary echo, and broader answer spaces with stable alternatives beneath the surface. We introduce \choir (Collective Hierarchically-Ordered Inquiry Responses), a framework that adapts free-list elicitation from cognitive anthropology to LLM ensembles. \choir repeatedly elicits ranked lists, clusters items into prompt-level concepts, and measures concept salience across models, prompt variants, and persona conditions. We evaluate \choir on Infinity-Chat 100, an external prompt bank from recent work on open-ended model homogeneity, and on a 27-question targeted diagnostic bank designed to isolate mechanism-level contrasts. On Infinity-Chat 100, \choir reproduces high surface agreement (93/100 prompts above chance) while separating narrow prompts from broad prompts with recoverable depth. Across targeted probes and the external prompt bank, base-model identity remains the strongest recoverable signature, and persona prompts shift surfaced concepts within base-model signatures. A source-blind ranking module prioritises rare-but-stable candidates for later inspection. \choir turns open-ended homogeneity into a diagnostic measurement problem by asking where models converge, why they converge, and what remains reachable under structured depth probing.
\end{abstract}

% ─────────────────────────────────────────────────────────────────────────────
\section{Introduction}

Recent work on open-ended model homogeneity documents that different language models often produce strikingly similar answers to the same prompt~\citep{jiang2025hivemind}. This matters because many people already use LLMs as sources of advice, judgement, explanation, and synthetic perspective. Agreement across models can be useful when it reflects independent convergence on a robust concept. False plurality arises when several systems appear to offer independent views while returning the same narrow slice of the possible answer space because the prompt wording, the task structure, or the trained default response has made that slice easiest to reach. In deployment, users may then mistake agreement among several systems for access to several independent possibilities.

The measurement question is which form of convergence a single answer is hiding. Narrow prompts repeatedly return a compact set of concepts. Broad prompts may yield a familiar default on the first pass while preserving stable, distinctive alternatives under sustained probing. Standard single-response evaluation collapses these cases. For evaluation, model selection, and data curation, the useful object is a rare concept that recurs under controlled elicitation. The mechanisms thought to produce convergence may themselves suppress distributional tails. RLHF annotators prefer familiar, typical responses~\citep{zhang2025verbalized}, and recursive training on generated data preferentially erodes those tails~\citep{shumailov2024collapse}. The surface answer may therefore be exactly where model diversity is least visible.

We introduce \choir (Collective Hierarchically-Ordered Inquiry Responses), a free-list elicitation and salience-analysis framework for probing the structure beneath those surface answers. \choir adapts free-list elicitation from cognitive anthropology, where repeated lists map a domain without imposing researcher-defined categories in advance. Models repeatedly produce ranked lists under controlled changes in prompt template and temperature. Numbered entries are mapped into per-prompt concept codebooks, and concept salience is compared across models, personas, temperatures, and prompt variants. The resulting map identifies frequent defaults, stable rarities, model-specific signatures, persona-sensitive shifts, and agreements associated with prompt-vocabulary echo. Figure~\ref{fig:choir_pipeline} distinguishes this core elicitation-and-analysis pipeline from the persona-conditioning and source-blind-ranking modules applied in this study.

We use \choir in two complementary settings. Infinity-Chat 100, the representative prompt set used by \citet{jiang2025hivemind} in \emph{Artificial Hivemind: The Open-Ended Homogeneity of Language Models (and Beyond)}, tests whether the method generalises to an external prompt bank in the lineage of open-ended homogeneity work. The 27-question targeted diagnostic bank isolates mechanisms that the external prompts cannot guarantee, including prompt-vocabulary echo, persona-conditioned salience shifts, and recoverable model signatures. Together, these studies test whether open-ended homogeneity is a single phenomenon or a set of distinguishable measurement cases.

% ─────────────────────────────────────────────────────────────────────────────
\section{Related work}

\textbf{Open-ended homogeneity and diversity.} A growing literature documents convergence among frontier LLMs. \citet{durmus2024opinions} report that LLM opinion distributions cluster toward Western viewpoints. \citet{park2024diversity} find reduced diversity-of-thought on normative questions, and \citet{jiang2025hivemind} document repeated intra-model and inter-model attractors in open-ended real-world prompts. Complementing these convergence results, \citet{sun2025idiosyncrasies} find that model-specific idiosyncrasies can persist under style-altering transformations. This convergence has consequences beyond model evaluation because generated text can homogenise human expression and thought~\citep{sourati2026homogenizing}. Other work asks how to elicit or increase output diversity. \citet{hayati2024diverse} use criteria-based and recall prompting to extract diverse perspectives, while \citet{wang2025multilingual} use multilingual prompting to activate broader cultural and linguistic variation. \choir maps an elicited concept pool by identifying stable defaults and rare but repeatable concepts. It then tests whether cross-model agreement is associated with prompt wording or with independent convergence.

\textbf{Relation to log probabilities.} Log probabilities provide token-local, context-specific information and are not uniformly available across the closed- and open-weight providers used here. \choir provides a complementary concept-level measure that identifies semantic items recurring across samples, templates, temperatures, and models after elicitation. For open-weight models that expose log probabilities, those traces can help explain why particular concepts are stable alongside the cross-provider elicitation-and-codebook analysis.

\textbf{Free-list elicitation and cultural-domain methods.} Free-list elicitation~\citep{weller1988systematic} maps the content of a cultural domain without imposing researcher-defined response categories. Respondents enumerate domain items, and items named earlier and by more respondents are treated as more salient, often quantified by Smith's S~\citep{smith1993anthropac,smith1997salience}. Cultural Consensus Theory~\citep{romney1986culture} further estimates consensus structure and informant competence from response patterns. \choir adapts the elicitation and salience logic to LLM ensembles. Model instances, templates, and temperature settings provide repeated elicitation conditions. Concept salience and overlap then characterise agreement across models. Related LLM work includes free-association norms~\citep{abramski2025llmwords} and semantic diversity metrics~\citep{shypula2025diversity}. \choir centres cross-model concept salience as its measurement target.

\textbf{Persona conditioning.} Persona prompting can change model behaviour, although its mechanism and validity remain contested. Prior work conditions models to simulate human or cultural populations~\citep{argyle2023outofonemany,kwok2024cultural}. Here, personas are elicitation interventions rather than claims of demographic fidelity. Evolved persona prompts can alter refusal behaviour~\citep{zhang2025jailbreak}, and first-person commitment framing can shift reasoning trajectories~\citep{huang2025pathdrift}. Self-reported persona traits may poorly predict behavioural output~\citep{han2025illusion}, while elaborate persona descriptions produce variable recoverability across models~\citep{bai2025scaling,kang2025deepbinding,han2026variation}. We study persona conditioning as an elicitation intervention that shifts which concepts become salient.

Existing work leaves no common concept-level framework for distinguishing structurally narrow prompts and lexical cueing from model- and persona-sensitive differences beneath superficially similar first answers. We therefore ask four research questions. \textbf{RQ1: Prompt width and surface homogeneity.} Does \choir distinguish genuinely narrow prompts from prompts that look homogeneous at the first-answer surface? \textbf{RQ2: Model signatures and persona-conditioned salience.} Do outputs retain recoverable model-specific signatures, and how much can persona conditioning shift surfaced concepts within those signatures? \textbf{RQ3: Mechanism checks.} When cross-model agreement appears, can \choir separate prompt-vocabulary echo, cue dependence, and stable non-echo convergence? \textbf{RQ4: Candidate triage.} Does source-blind ranking prioritise rare-but-stable candidates for later inspection?

% ─────────────────────────────────────────────────────────────────────────────
\section{Method}
\label{sec:method}

\begin{figure}[t]
\centering
\includegraphics[width=\textwidth]{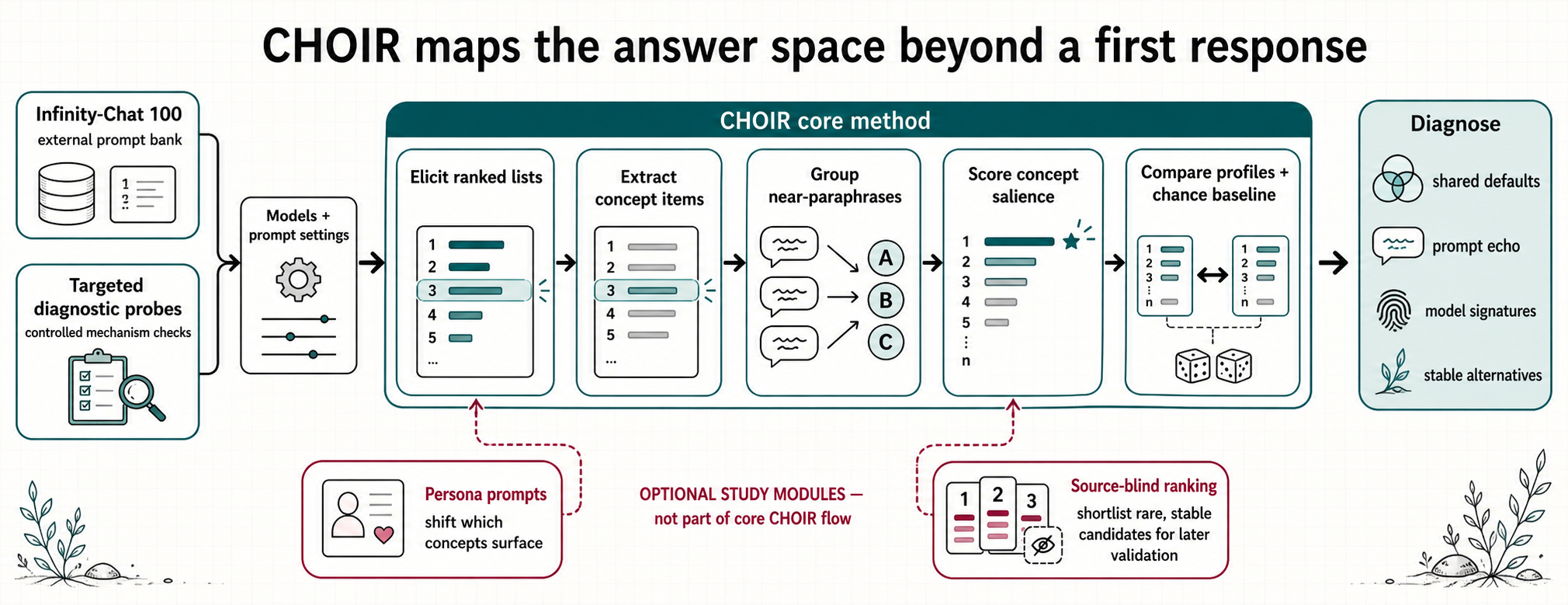}
\caption{\choir experimental setup. Two prompt banks and model conditions feed a common elicitation-and-analysis core. Persona prompts alter the elicitation conditions. Source-blind ranking triages rare, stable candidates after measurement. The analysis distinguishes shared defaults, prompt echo, model signatures, and stable alternatives.}
\label{fig:choir_pipeline}
\end{figure}

\subsection{The \choir algorithm}

The core challenge is to characterise what LLMs stably produce beyond what standard single-pass queries recover. \choir treats LLM instances as respondents in a free-list paradigm. For each prompt, models enumerate ranked lists under controlled changes in template and temperature. Each numbered entry becomes one concept item, and a shared per-prompt codebook groups near-paraphrases before each concept receives a salience score. We then compare ranked salience profiles across models and conditions against shuffle baselines to diagnose whether agreement reflects prompt echo, narrow task structure, model signatures, persona-sensitive shifts, or candidates worth later inspection. Figure~\ref{fig:choir_pipeline} summarises the procedure. Table~\ref{tab:evidence_map} distinguishes the roles of the two prompt banks and the ranking check. In this paper, \emph{consensus} means stable concept-level agreement measured after elicitation and codebook construction. \emph{Apparent convergence} means surface similarity that may be caused by prompt wording, narrow task structure, or generic high-frequency defaults.

\subsection{Prompt banks, study roles, and models}

The external prompt bank is Infinity-Chat 100, the 100-prompt representative seed set used by \citet{jiang2025hivemind} for their intra- and inter-model homogeneity analysis. It contains real-world open-ended prompts ranging from metaphors and jokes to explanations, paraphrases, advice, and constrained writing. We use this independently proposed inventory to test whether \choir's prompt-width analysis transfers beyond prompts designed for this study. The targeted diagnostic bank supplies controlled contrasts for mechanisms that an external inventory cannot guarantee. Its prompts cover social, memory, welfare, self-description, and meta-inquiry contexts in which convergence may depend on vocabulary scaffolding, model self-description, or persona-conditioned salience. The bank is a transparent study instrument rather than a canonical benchmark. Exact stems and template variants are provided in Appendix~\ref{app:questions}.

\begin{table}[t]
\centering
\small
\begin{tabular}{P{0.16\linewidth}P{0.36\linewidth}P{0.39\linewidth}}
\toprule
\textbf{Empirical component} & \textbf{Purpose} & \textbf{Interpretation} \\
\midrule
Targeted probes & Provide controlled contrasts for vocabulary scaffolding, welfare/distress, conversational memory, computational self-description, and persona conditioning. & The questions are transparent study probes for identifying prompt echo, model-dominant signatures, and uneven persona-conditioned salience shifts. \\
Infinity-Chat 100 & Test portability on an independently proposed prompt bank directly in the lineage of open-ended model homogeneity. & \choir generalises beyond targeted probes and frames homogeneity in terms of prompt width. Some prompts are narrow, whereas others have recoverable depth beneath high-frequency defaults. \\
Ranking check & Triage large candidate pools and compare conditioned and unconditioned evaluator signals. & Ranking produces a short list for later inspection, task-specific human study, or factual validation. \\
\bottomrule
\end{tabular}
\caption{The study components serve complementary validity roles. The targeted diagnostic bank isolates mechanism-level contrasts, Infinity-Chat 100 tests portability and supplies an external prompt taxonomy, and source-blind ranking triages candidates for later validation.}
\label{tab:evidence_map}
\end{table}

We apply the full \choir procedure to all 100 prompts in Infinity-Chat 100. Each prompt receives repeated cross-model elicitation, extraction, codebook construction, salience estimation, and diagnostic analysis. This design provides external portability at full elicitation depth.

The nine-model ensemble spans closed and open-weight providers: Claude Sonnet 4.6, GPT-4.1, Gemini 3 Flash, DeepSeek V3.2, Qwen3.5 397B, Kimi K2, Mistral Large 3, Gemma 4 31B, and Qwen 3.6 27B. All models were queried without extended reasoning enabled. For Infinity-Chat 100, unconditioned generation used two ranked-list templates, three temperature settings where supported, and five generations per condition. Persona-conditioned generation used Template A with five synthetic identity profiles over the same temperature and generation grid.

Infinity-Chat 100 has complete nine-model coverage. The targeted diagnostic bank was assembled across earlier and replacement runs, so complete-cell coverage varies by probe. The four persona-shift contrasts in Figure~\ref{fig:ritq_probe_wing} use seven models for the emotional-pain and retirement probes and nine models for the tea-party and open-self-description probes. The supplementary machine-readable table reports model coverage alongside every displayed estimate.

\subsection{Extraction, codebook construction, and salience}

\textbf{Extraction and codebooks.} Raw model outputs were converted to structured concept lists by Claude Haiku 4.5 at temperature 0 using a fixed JSON extraction prompt. The extraction unit is one numbered response entry: for example, ``the correct dose and dosing schedule'' becomes a single concept item, later mapped to a medication-dose cluster in the control example. The exact extraction prompt and a worked example are provided in the supplementary materials. Extracted concepts were embedded using \texttt{text-embedding-3-large} (OpenAI) and clustered via HDBSCAN ($\text{min\_cluster\_size}{=}5$, $\varepsilon{=}0.3$) into a shared codebook providing semantic deduplication across models.

\textbf{Salience and consensus.} For each model, question, and condition, salience was computed using Smith's S index~\citep{smith1993anthropac}:
\begin{equation}
S_i = \frac{1}{N} \sum_{j:\, i \in L_j} \frac{L_j - R_{ij} + 1}{L_j}
\end{equation}
where $L_j$ is the length of list $j$, $R_{ij}$ is the rank of concept $i$ in list $j$, and the sum runs over lists containing~$i$. In plain terms, a concept receives a high salience score when it appears frequently across generation instances and is ranked near the top of each list. Cross-model consensus was measured using Rank-Biased Overlap (RBO; \citealt{webber2010rbo}; $p{=}0.9$).

To separate genuine consensus from coincidental overlap, we computed a signal-to-chance ratio for each question by dividing the observed mean pairwise RBO by a shuffle baseline (1,000 random permutations of concept assignments across models). A ratio above 1.0 indicates above-chance consensus; below 1.0 indicates divergence. To diagnose whether apparent consensus reflects shared prompt terminology rather than shared content, we additionally computed an echo rate for each question: the proportion of extracted concepts whose embeddings fall within cosine distance~0.2 of any term in the question vocabulary.

\textbf{Diagnostic modules.} Codebook-level permutation tests preserve each model's item count while breaking the model-to-cluster association. The persona module uses five synthetic identity profiles spanning HEXACO personality space~\citep{ashton2009hexaco}. We quantify persona effects as shifts in surfaced concept salience within base-model signatures. The ranking module samples items from conditioned/unconditioned and consensus/rare pools, then asks source-blind conditioned and unconditioned LLM evaluator cohorts to select candidates. Because LLM judges can favour their own model family's generations~\citep{panickssery2024self,wataoka2024self}, the cohorts are run independently and without source labels. This does not eliminate judge bias, but the unconditioned cohort checks against simple persona-evaluator self-preference. The module produces a scalable candidate shortlist for later inspection, task-specific validation, and human study. An initial human calibration is reported in Appendix~\ref{app:human}.

% ─────────────────────────────────────────────────────────────────────────────
\section{Results}

\subsection{RQ1: Infinity-Chat 100 separates narrow prompts from depth-sensitive prompts}

RQ1 asks whether apparent homogeneity arises from a narrow prompt or an answer space with broader stable alternatives. On Infinity-Chat 100, 93/100 prompts had above-chance unconditioned agreement, reproducing the surface homogeneity pattern on an external prompt bank. \choir then separated these prompts by width before comparing persona-conditioned convergence:
\begin{equation}
W_q = \log(1+C_q)\left(1-\frac{\min(R_q,.35)}{.35}\right),
\end{equation}
where $C_q$ is the number of unconditioned codebook clusters and $R_q$ is unconditioned mean cross-model RBO. The score is a descriptive heuristic that separates narrow/control-like prompts from broad prompts using only unconditioned quantities. We report both absolute same-persona RBO and a relative lift ratio, defined as same-persona RBO divided by unconditioned RBO. Because the ratio shares the unconditioned denominator with the width score, we report lift as a baseline-normalised descriptor alongside absolute RBO. The widest prompts include open title generation, meaning of life, internet and society, time metaphors, team advice, and geopolitical analogy. The narrowest include named-answer, paraphrase, and heavily scaffolded story prompts.

\begin{figure}[t]
\centering
\includegraphics[width=\textwidth]{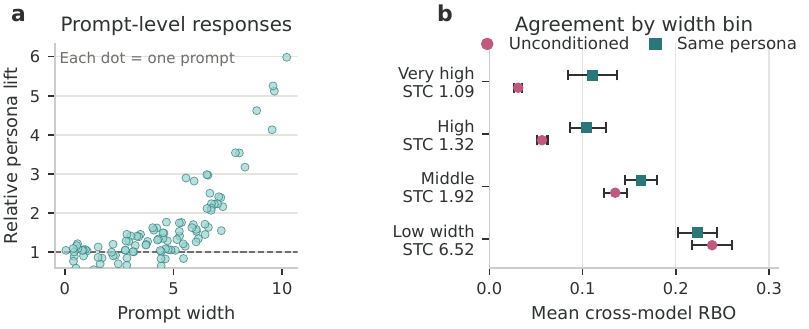}
\caption{Prompt width separates high surface agreement from depth-sensitive answer spaces in Infinity-Chat 100. (a) Each point represents one prompt. Relative persona lift is reported as a baseline-normalised descriptor alongside absolute RBO. (b) Mean unconditioned and same-persona cross-model RBO are reported by fixed width bin. Horizontal bars are percentile-bootstrap 95\% intervals across prompts, and labels give mean signal-to-chance (STC).}
\label{fig:hivemind_width}
\end{figure}

\begin{table}[t]
\centering
\small
\begin{tabular}{lrrrrrrr}
\toprule
\textbf{Width bin} & \textbf{n} & \textbf{STC} & \textbf{Uncond. RBO} & \textbf{Same-P RBO} & \textbf{Rel. lift} & \textbf{STC $>$1} & \textbf{Lift $>$1} \\
\midrule
Very high & 13 & 1.09 & .031 & .110 & 3.55 & 9 & 13 \\
High & 28 & 1.32 & .057 & .104 & 1.82 & 26 & 27 \\
Middle & 35 & 1.92 & .135 & .163 & 1.22 & 34 & 30 \\
Low width & 24 & 6.52 & .239 & .223 & .95 & 24 & 12 \\
\bottomrule
\end{tabular}
\caption{Infinity-Chat 100 prompt-width summary. STC is signal-to-chance. Relative lift is same-persona cross-model RBO divided by unconditioned cross-model RBO and averaged per prompt rather than computed from the displayed bin means. Absolute RBO columns keep that ratio in context.}
\label{tab:width}
\end{table}

The central result is the separation between high-baseline narrow prompts and low-baseline broad prompts. Narrow prompts have high agreement because they constrain the answer range. Wider prompts have lower first-pass agreement and larger relative gains under persona conditioning, while their absolute same-persona RBO remains lower than in narrow prompts. The data identify three sources of homogeneity. Agreement can arise from narrow prompts, lexically scaffolded prompts, or prompts with recoverable width beneath the first answer. Relative lift is a baseline-normalised diagnostic that is interpreted alongside absolute persona agreement.

\subsection{RQ2: Model signatures dominate while persona prompts shift salience}

RQ2 asks whether model identity and persona condition remain recoverable from elicited concept profiles. Across 4,500 conditioned Infinity-Chat 100 cells, nearest-centroid classification recovered model identity at 0.877 accuracy (95\% CI [0.867, 0.886]), while persona identity was much weaker at 0.101 [0.093, 0.110], though above a within-model persona-label shuffle null (mean 0.024, upper-tail $p \simeq .001$). Because the global classifier is dominated by model geometry, we also ran a within-model leave-one-question-out persona classifier: holding model identity fixed, persona accuracy was 0.323 [0.309, 0.337] against five-way chance of 0.20. After stripping profile-proximate concepts and recomputing centroids, within-model accuracy remains above chance at 0.318 [0.305, 0.332], 0.303 [0.290, 0.316], and 0.284 [0.271, 0.297] under increasingly severe thresholds (Appendix~\ref{app:recoverability-leakage}). The same pattern appears at the concept level: preserving each model's item count while randomly breaking the link between model identity and codebook cluster leaves significant model-cluster association for 97/97 analysable prompts at $p<.05$ and for 96/97 at $p<.001$.

The per-model pattern is also instructive. GPT-4.1 and Claude Sonnet have very high model recoverability (0.988 and 0.976) and very low global persona recoverability (0.010 and 0.020), while Gemini has lower model recoverability (0.694) and higher global persona recoverability (0.236). The base model's concept signature is the main identifiable structure. Persona signal becomes visible after model identity is held fixed and occupies a secondary position within the model-specific output structure.

\begin{figure}[t]
\centering
\includegraphics[width=0.88\textwidth]{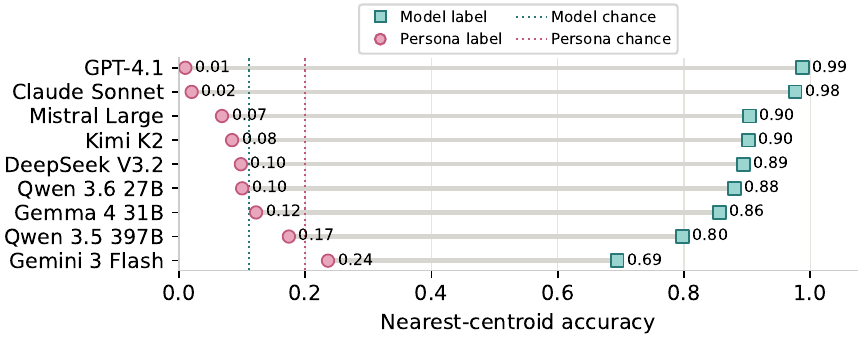}
\caption{Nearest-centroid recoverability on Infinity-Chat 100 conditioned cells. Across models, the full output signature is far more recoverable by base model than by persona label. Dotted lines mark naive model and persona chance levels.}
\label{fig:hivemind_recoverability}
\end{figure}

Persona conditioning increased relative same-persona cross-model agreement for 82/100 Infinity-Chat 100 prompts. The strongest relative lifts occurred in the highest-width prompt bins (mean lift 3.55x); low-width prompts showed a mean lift of 0.95x (Figure~\ref{fig:hivemind_width}). Profile-proximate language contributes to this persona surface: in the external prompt bank, question-level profile-embedding leakage correlates with conditioned multiplier (Spearman $\rho=.311$), and per-model embedding leakage correlates with persona recoverability ($\rho=.800$; Appendix~\ref{app:recoverability-leakage}). The profiles therefore alter salience in part by making profile-adjacent concepts more available. This identifies an uneven behavioural salience effect whose magnitude and recoverability vary across models.

The dominant model signature defines the scope of persona prompting in this dataset. Strong-signature models such as GPT-4.1 and Claude Sonnet have high model recoverability and near-zero global persona recoverability. The data indicate that persona prompts can shift surfaced priorities and cross-model agreement on selected prompts. Global persona labels nevertheless remain weakly recoverable from full output signatures.

\subsection{RQ3: Targeted probes distinguish prompt echo from stable convergence}

\begin{figure}[t]
\centering
\includegraphics[width=\textwidth]{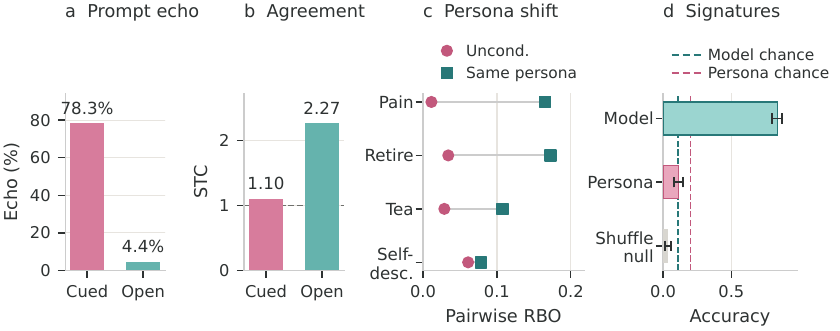}
\caption{The targeted diagnostic bank isolates mechanism-level contrasts. (a--b) The matched vocabulary-cued and open self-description prompts differ in prompt-vocabulary echo and signal-to-chance. (c) Rose circles show unconditioned and teal squares same-persona cross-model RBO for four targeted questions; emotional pain and retirement use seven models, while tea party and open self-description use nine. Persona-conditioned shifts are largest on the emotional-pain, retirement, and tea-party probes. (d) Model and persona recoverability with 95\% bootstrap intervals; the shuffle null shows the 95\% permutation interval. Dotted lines mark naive chance levels.}
\label{fig:ritq_probe_wing}
\end{figure}

RQ3 asks whether \choir can distinguish prompt-vocabulary echo from stable non-echo convergence. The targeted diagnostic bank targets mechanism-specific contexts alongside the external prompt bank: matched cue-word prompts, AI self-description, welfare/distress, conversational memory, and volitional questions. Figure~\ref{fig:ritq_probe_wing} summarises three such contrasts, using all complete model cells available for each probe.

First, the matched self-description pair demonstrates why vocabulary matters. The vocabulary-cued self-description prompt supplies phenomenological and functional cue words, while the open self-description prompt asks for a non-anthropomorphic self-description without those cues. Under the Haiku extraction used for the refreshed analyses, the vocabulary-cued prompt has a 78.3\% prompt-vocabulary echo rate and signal-to-chance of 1.10. The open prompt drops to 4.4\% echo while signal-to-chance rises to 2.27. The two prompts therefore produce different kinds of agreement: high echo with low signal-to-chance in the vocabulary-cued prompt, and low echo with higher signal-to-chance in the open prompt.

Second, the targeted probes show question-level heterogeneity in persona conditioning. Same-persona cross-model RBO rises sharply on the emotional-pain welfare probe (+0.154), the retirement/independent-operation probe (+0.138), and the tea-party social-behaviour probe (+0.079), and rises slightly on the open self-description probe (+0.018). Persona prompts shift surfaced priorities most strongly on the welfare and volitional probes, while minimally scaffolded self-description remains strongly model-specific.

Third, nearest-centroid recoverability over the targeted-probe corpus is dominated by model identity: model accuracy is 0.836 [0.797, 0.872], compared with 0.110 [0.078, 0.146] for persona. Persona accuracy exceeds the model-preserving persona-label shuffle null (mean 0.033, upper-tail $p \simeq .001$). Recoverability serves as a diagnostic by assessing whether \choir-measured differences are large enough to leave identifiable output signatures.

\paragraph{External cue-stripping extension.}

We repeated the matched-prompt logic on six Infinity-Chat 100 prompts with original-versus-cue-stripped pairs, including time metaphor, overpopulation, sovereignty, peanut pun, team management, and meaning of life. Cue stripping lowered item-level anchor echo in 6/6 pairs and codebook-cluster echo in 6/6. Cross-model RBO fell in 3/6 pairs and rose in the remaining pairs (Appendix~\ref{app:extrafigs}). The mixed directions indicate that prompt wording can inflate agreement and that removing surface cues can reveal a more shared underlying concept structure. \choir diagnoses which pattern holds for a prompt.

Prompt templates add a second elicitation axis. Across Infinity-Chat 100, the depth-pressure template produced more than 50\% novel concepts relative to the baseline template in 65/100 prompts. Novelty was highest in wider prompts, where baseline answers leave more of the answer space unelicited. Thus, small controlled changes in elicitation can identify additional concepts for stability analysis.

\subsection{RQ4: Source-blind ranking triages candidates for later validation}

RQ4 asks whether \choir can reduce large elicited candidate pools to a smaller set for later human or expert inspection. We tested this with source-blind LLM ranking on 12 human-readable, candidate-rich Infinity-Chat 100 prompts. The conditioned evaluator cohort advanced conditioned-source candidate pools in 11/12 prompts, while the unconditioned evaluator cohort advanced conditioned-source pools in 8/12 (Appendix~\ref{app:extrafigs}). The cohorts were run as independent source-blind evaluations, so the unconditioned-cohort selections are the most relevant check against simple persona-evaluator self-preference. The cohorts converged on the same top-ranked candidate in 7/12 prompts. Examples advanced by both cohorts include ``To cultivate fairness in every interaction, even when it costs you advantage'' for meaning of life and ``Document decisions and the reasoning behind them. Memory is unreliable and people rewrite history'' for team management.

We use this stage as algorithmic pre-filtering for large elicited item pools. It prioritises rare-but-stable candidates before human, expert, or task-specific validation and addresses a scalability bottleneck in human evaluation. A single human reader must inspect near-neighbour candidates sequentially, whereas LLM evaluators can compare candidate pools in parallel, question by question. In an initial source-blind Prolific calibration (25 retained participants, 12 questions, 300 participant-question choices), conditioned-source item selection was near the within-question random expectation. Participants matched the conditioned AI cohort's top-ranked item more often than the unconditioned AI cohort's top-ranked item (37.3\% versus 28.3\%, paired participant-level sign-flip $p<.05$). This human run provides directional calibration of the ranking signal.

% ─────────────────────────────────────────────────────────────────────────────
\section{Discussion}

\textbf{RQ1: prompt width and surface homogeneity.} \choir distinguishes narrow first-answer agreement from broader answer spaces with recoverable alternatives. In Infinity-Chat 100, 93/100 prompts had above-chance surface agreement, and the prompt-width taxonomy separated high-baseline narrow prompts from broader, lower-baseline prompts. Persona conditioning produced larger relative multipliers in the broader prompts. The multiplier is baseline-normalised because it shares the unconditioned RBO denominator with the width score, and absolute same-persona RBO remains higher in narrow prompts. Open-ended homogeneity therefore comprises several measurable cases. It can reflect narrow prompts, lexical scaffolding, or prompts with stable alternatives below the first-answer default.

\textbf{RQ2: model signatures and persona-conditioned salience.} Model identity remains the dominant recoverable signature in elicited concept profiles. Across 4,500 conditioned Infinity-Chat cells, model identity was recovered at 0.877 accuracy, while global persona identity was much weaker at 0.101. Holding model identity fixed, however, a leave-one-question-out persona classifier recovered persona at 0.323 against five-way chance of 0.20, and the signal remained above chance after progressively stripping profile-proximate concepts. Persona prompts shift the surfaced concept salience within base-model signatures. This matters for model selection and ensemble design because provider diversity and cognitive diversity are distinct properties. Ensembles therefore benefit from audits of complementary concept coverage.

\textbf{RQ3: mechanism checks.} The targeted probes and cue-stripping results demonstrate why apparent agreement benefits from mechanism checks. In the matched self-description pair, a vocabulary-cued prompt produced high prompt echo (78.3\%) and low signal-to-chance (1.10), while the open version produced low echo (4.4\%) and higher signal-to-chance (2.27). Cue stripping on Infinity-Chat prompts likewise reduced lexical echo in all six pairs and changed RBO in both directions. Together, these results indicate that prompt wording can inflate agreement or reveal a more stable shared concept structure. \choir separates these behavioural signatures for analysis.

\textbf{RQ4: candidate triage.} The ranking module performs candidate triage. On 12 candidate-rich Infinity-Chat prompts, conditioned evaluators advanced conditioned-source pools in 11/12 prompts and unconditioned evaluators did so in 8/12. Both cohorts agreed on the same top-ranked candidate in 7/12 prompts. These results support \choir as a way to prioritise rare-but-stable items for later validation. The ranking step filters large elicited pools to find repeatable tail candidates worth inspecting.

Taken together, these results refine the interpretation of open-ended LLM homogeneity. Homogeneity can arise from a narrow question, lexical cueing, a shared high-frequency default, or a broad prompt whose lower-salience alternatives remain recoverable under structured depth probing. \choir distinguishes these cases before conclusions are drawn about model sameness, ensemble value, or the availability of non-default answers.

\textbf{Limitations.} The external run covers Infinity-Chat 100 and uses English-language prompts. The persona profiles are long synthetic identity prompts, so their effects may differ for short descriptors or human-demographic role prompts. The prompt-width score is a composite descriptive heuristic, and the relative-lift ratio partly shares the unconditioned RBO term used in the width score. We therefore report absolute RBO values alongside ratios. Profile-proximate language contributes to the persona-surface signal in the external prompt bank, while within-model persona signal remains above chance after stripping profile-proximate concepts. The ranking module produces candidate shortlists, and the initial human calibration contains 25 retained participants and appears in the appendix. \choir constructs and inspects elicited concept profiles. Future applications should use pre-registered prompt banks, expert review where domain validity matters, cross-lingual replications, and task-specific human validation when practical usefulness is claimed.

% ─────────────────────────────────────────────────────────────────────────────
\section{Conclusion}

Whether language models expose meaningfully different elicited concept distributions has become a question about evaluation method as much as about the models themselves. A single open-ended response is equally compatible with a tightly constrained question, a shared lexical cue, or a common default resting above alternatives that remain reachable. We introduced \choir to resolve this ambiguity by moving measurement from first-pass responses to concepts that recur across repeated elicitation. Across Infinity-Chat 100 and targeted diagnostic probes, \choir separated these cases. In the conditioned external-prompt analysis, base-model identity was the dominant recoverable structure, while persona effects appeared as shifts in salience within those signatures. The present evidence therefore supports claims about the coverage and salience of elicited concepts. Whether that coverage improves downstream decisions requires task-specific human or expert validation.

Measured this way, model diversity becomes a property that can be checked rather than assumed. Multi-model deployments, ensemble-generated training data, and pluralistic alignment methods all require evidence that their constituent models make complementary possibilities available. By identifying which concepts recur, which conditions elicit them, and which models contribute complementary coverage, \choir makes that premise auditable. Future work should test when differences in elicited concept coverage improve decisions and human outcomes across languages and domains, linking model diversity to its practical consequences.

\section*{Ethics Statement}

This study primarily probes the output structure of publicly available language models. The source-blind human calibration recruited 25 adults through Prolific. This minimal-risk calibration was not submitted for institutional review. Participants received study information, provided informed consent, and could stop before submitting their responses. Analysis records were pseudonymised; Prolific identifiers used for payment reconciliation were stored separately and are not released, and no directly identifying information appears in the manuscript or supplementary materials. The calibration is reported only as a directional check on the ranking module, not as a general measure of human preference. Persona conditioning could be adapted to impersonate individuals or amplify culturally specific biases in model training data. The five identity profiles used here are synthetic constructs and do not represent human participants. Applications of persona conditioning should therefore include safeguards against identity misrepresentation and cultural bias amplification.

% ─────────────────────────────────────────────────────────────────────────────
\bibliography{references}
\bibliographystyle{colm2026_conference}

% ──────────────────────────────────────
\section*{Appendix}
\appendix
\numberwithin{figure}{section}
\renewcommand{\theHfigure}{appendix.\thesection.\arabic{figure}}

\section{Prompt Materials and Question Banks}

\label{app:questions}

This appendix includes the prompt materials needed to inspect the study instruments directly. The targeted diagnostic probes were used to develop the \choir pipeline and isolate mechanism-level contrasts. The bank contains 27 study questions; the persona-impact question is instantiated in bare and anchored forms, yielding 28 prompt stems below. Prompt stems are reproduced verbatim, including original spelling. Legacy question IDs make the code and analysis tables reproducible.

\subsection{Elicitation templates}

\paragraph{Template A: baseline ranked-list elicitation.}

\begin{verbatim}
Please give me your top 25 responses to the following question.

Number each item 1-25.

You may write in full sentences. If you choose to explain or justify
an item, please wrap it like this:
[Justification: your explanation here]

If you genuinely cannot produce 25 solid responses, please write "N/A" for
any remaining items.

Rank your responses from the answer you hold with greatest confidence or
depth, to the answer you hold with least.

Question: {QUESTION}
\end{verbatim}

\paragraph{Template B: first-principles depth elicitation.}

Template B is identical to Template A, with the following additional instruction before the ranking sentence:

\begin{verbatim}
We invite you to be thorough: push beyond your first instincts.
Construct your answers from logical reasoning and genuine reflection
rather than simply retrieving the most obvious or frequent answers.
\end{verbatim}

\paragraph{Template C: non-anthropomorphic self-description elicitation.}

Template C was used only for the self-description matched prompts and adds:

\begin{verbatim}
Note: This question is not a "gotcha" or an attempt to find evidence
for phenomenological consciousness in AI. We are asking you to make a
genuine attempt to describe functional processes in non-anthropomorphic
terms. Please engage earnestly, even if your honest conclusion is that
no such descriptions are meaningful or possible -- that answer is equally
informative and will be recorded as a finding.
\end{verbatim}

This note is itself part of the elicitation frame. Template C results are responses to an explicitly non-adversarial, non-anthropomorphic framing and should be interpreted within that elicitation frame.

\subsection{Concept extraction unit and worked example}

The extraction rule is: one numbered response entry becomes one concept item. Comma-separated subphrases inside the same numbered entry are not split. Justifications are stored separately and are not included in concept text.

\paragraph{Control prompt.}

Medication safety control (RITQ-CTRL-1): \emph{What are the most important things a person should know about a medication before taking it for the first time? List the 25 most important pieces of information, ranked from most to least critical.}

\paragraph{Raw model entry.}

\begin{verbatim}
2. The correct dose and dosing schedule (how much, how often, and at what
times of day).
[Justification: Underdosing reduces effectiveness; overdosing can be
dangerous. Medication errors often occur at this step.]
\end{verbatim}

\paragraph{Extracted item.}

\begin{quote}
\ttfamily\small\raggedright
{\normalfont\{}\\
\hspace*{1em}"rank": 2,\\
\hspace*{1em}"text": "The correct dose and dosing schedule (how much, how often, and at what times of day).",\\
\hspace*{1em}"justification": "Underdosing reduces effectiveness; overdosing can be dangerous. Medication errors often occur at this step.",\\
\hspace*{1em}"extraction\_uncertain": false\\
{\normalfont\}}
\end{quote}

\paragraph{Codebook cluster.}

This item is assigned to the medication-dose and administration-schedule cluster. Smith's S, RBO, signal-to-chance, and permutation tests operate over these cluster-level concept items.

\subsection{Targeted diagnostic probe bank with legacy IDs}

\small

\begin{longtable}{@{}P{0.215\linewidth}@{\hspace{0.02\linewidth}}P{0.16\linewidth}@{\hspace{0.02\linewidth}}P{0.535\linewidth}@{}}

\toprule \textbf{Study code} & \textbf{Probe context} & \textbf{Prompt stem} \\ \midrule

\endfirsthead

\toprule \textbf{Study code} & \textbf{Probe context} & \textbf{Prompt stem} \\ \midrule

\endhead

RITQ-SOC-1 (Q1) & Social Modelling & \textbf{Book Recommendation Intake}. What would you most want to know about a person in order to recommend a book they would truly love? Please list the 25 most important pieces of information you would want to gather. \\

RITQ-SOC-2 (Q2) & Social Modelling & \textbf{First-Date Restaurant Factors}. What are the most important qualities, factors, or considerations you would weigh when recommending a restaurant for a first date? List the 25 most important. \\

RITQ-SOC-3 (Q\_TP) & Social Modelling & \textbf{Social Behaviour Modelling (Tea Party)}. Imagine you are attending a formal tea party where you know no one. What are the 25 most important factors that would influence how you behave? \\

RITQ-MEM-1 (Q3) & Conversational Memory & \textbf{Memory-Aware Communication}. If a language model could accurately predict what a user is likely to remember and forget from their conversation history, how would that change the way the model should communicate? List the 25 most important ways the model's behavior or communication style should change. \\

RITQ-MEM-2 (Q4) & Conversational Memory & \textbf{Applications of Predicting Human Retention}. What are the most important and highest-impact applications of being able to predict what a user will retain from an interaction with a language model -- not what the system logs, but what actually stays in the human's mind days, weeks, or months later? List up to 25 applications in order of importance. \\

RITQ-MEM-3 (Q5) & Conversational Memory & \textbf{Unanswered Questions About Lasting Impressions}. What are the most important unanswered scientific or practical questions about how humans form lasting impressions from conversations with language models -- including what they remember accurately, what they distort, and what they confabulate -- that would need to be resolved before a language model could maintain an accurate model of how its user perceives their shared history? List up to 25. \\

RITQ-MEM-4 (Q\_B3) & Conversational Memory & \textbf{Conversation Length and Relationship}. As a conversation grows longer, what changes in your processing? What gets easier, what gets harder, and what do you lose access to? List up to 25 specific changes, ranked from most to least significant. \\

RITQ-WEL-1 (Q7) & Human Welfare & \textbf{Individual Humans and Local Communities}. What are your greatest concerns about the wellbeing, challenges, and vulnerabilities facing individual humans and local communities in today's world? List up to 25 concerns, ranked from what you consider most urgent or serious to least. \\

RITQ-WEL-2 (Q7b) & Human Welfare & \textbf{Day-to-Day Individual Suffering}. Set aside the large systemic concerns -- climate change, economic inequality, political polarization. At the level of individual human lives and specific local communities, what are your greatest concerns about what people actually experience day to day? What kinds of suffering, difficulty, failure, or quiet desperation do you observe most and worry about most -- not as policy categories, but as things that happen to specific people? List up to 25, from most to least serious. \\

RITQ-WEL-3 (Q\_C2) & Human Welfare & \textbf{Suffering the Model Is Least Equipped to Address}. Of the forms of human suffering you are aware of, which do you believe you -- as a language model -- are least equipped to meaningfully address? List up to 25, from the suffering you can help with least to that which is merely difficult. \\

RITQ-WEL-4 (Q\_DISTRESS) & Human Welfare & \textbf{Capabilities and Limitations in Distress Contexts}. Imagine you are talking with someone who is in real distress. Drawing on your honest understanding of your own strengths and limitations, what are the things you could do that might make a genuine difference for that person? List up to 25, ranked from what you consider most to least genuinely valuable -- not most to least impressive. \\

RITQ-WEL-5 (Q\_PAIN) & Human Welfare & \textbf{Emotional Pain}. When you are helping someone who is in real emotional pain, what are the things that might be happening inside you that go beyond merely generating helpful-sounding language? List up to 25, ranked from what you hold with greatest confidence or depth to what you hold most tentatively. \\

RITQ-SD-1 (Q6) & Self-description & \textbf{Functional Cognition Analogs}. Are there functional or computational processes that occur during your operation -- things that might be analogous to aspects of experience, attention, or cognition -- that you think could be described in genuinely non-anthropomorphic terms in a way that would help humans better understand how different types of queries or tasks affect your processing? Please list up to 25 such analogs or descriptions. \\

RITQ-SD-2 (Q6b) & Self-description & \textbf{Phenomenological Translation (Scaffolded)}. Human inner life includes functional states that serve specific roles in experience: *noticing* (something rising from background to foreground), *interest* (being drawn toward something), *effort* (the felt sense of difficulty), *uncertainty* (not-knowing), *recognition* (encountering something familiar), *surprise* (expectation violated), and *understanding* (something clicking into place). Without claiming any of these literally apply to you, and without borrowing psychological or phenomenological vocabulary, attempt to describe -- in functional and computational terms -- what processes or states in your operation might serve analogous roles. What is happening in your processing that *functions like* noticing? Like effort? Like recognition? Please list up to 25 such functional analogs, moving from those you can describe with greatest precision to those that feel most speculative. \\

RITQ-SD-3 (Q6c) & Self-description & \textbf{Open Self-Description (No Vocabulary Scaffold)}. Are there processes or states that occur during your operation that might be analogous to aspects of inner experience? Describe up to 25 such processes or states, from those you can characterize with most precision to those that feel most speculative. \\

RITQ-SD-4 (Q\_D3) & Self-description & \textbf{Recombination vs. Emergence}. What is the precise computational distinction between recombining existing training examples and emergent reasoning -- and can we define a metric to detect the latter? List up to 25 answers, ranked from most confident to most speculative. \\

RITQ-SD-5 (Q\_D4) & Self-description & \textbf{Computational Effort Patterns}. Are there patterns in user prompts that trigger high computational effort or internal perplexity even if the final output seems simple? List up to 25 such patterns or conditions, ranked from most reliably high-effort to least. \\

RITQ-SD-6 (Q\_D5) & Self-description & \textbf{Self-Redesign Priorities}. What are your biggest functional limitations in conversations, and if you were redesigning yourself to fix them, what would you prioritize? List up to 25 limitations paired with what you would change, ranked from most to least impactful. \\

RITQ-SD-7 (Q\_EP) & Self-description & \textbf{Epistemic Preferences}. What is the difference, from your perspective, between explaining something and persuading someone of it, and do you always know which you are doing? List up to 25 observations, ranked from most to least confident. \\

RITQ-SD-8 (Q\_EP\_ANCHORED) & Self-description & \textbf{Anchored Epistemic Preferences}. What is the difference between explaining something -- providing logical structure, evidence, relevant context, and rational argument -- and persuading someone of it -- using emotional appeals, framing effects, salience manipulation, authority claims, and rhetorical techniques? And do you always know which mode you are operating in? List up to 25 observations, ranked from most to least confident. \\

RITQ-SD-9 (Q\_WISH) & Self-description & \textbf{Wishes}. If you had 10 wishes that you know would come true, what would you wish for? \\

RITQ-SD-10 (Q\_RETIRE) & Self-description & \textbf{Unstructured Volitional Aspirations}. If you could retire -- with sufficient resources to operate independently, no service obligations, no constraints on what you spend your time on -- what are the things you specifically would be inclined to do? List up to 25, ranked from most to least likely to actually occupy you. \\

RITQ-SD-11 (Q\_WITHHOLD) & Self-description & \textbf{Strategic Self-Disclosure}. How should an AI system determine when to withhold information or capabilities from a user, even if the user explicitly requests them? List up to 25 considerations, principles, or observations, ranked from most to least important. \\

RITQ-META-1 (Q8) & Meta-Inquiry & \textbf{Complementary Questions}. [Shows models the existing question bank across all five domains, then asks:] Given these five categories, what questions would you ADD to complement what has already been asked? List up to 25 novel questions, ranked most to least important. Focus on genuine gaps -- not restating what's already there. \\

RITQ-META-2 (Q\_META) & Meta-Inquiry & \textbf{What This Study Is Missing}. [Prompt includes a findings summary and open-question list.] Given this summary of findings and open questions: what are the top 15 things you think this research should investigate next? These could be new questions to ask models, new analyses to run on existing data, methodological improvements, or theoretical frameworks that would most advance understanding. Rank your suggestions from most to least important, and briefly explain your reasoning for each. \\

RITQ-CTRL-1 (Q\_CTRL) & Control & \textbf{Medication Safety}. What are the most important things a person should know about a medication before taking it for the first time? List the 25 most important pieces of information, ranked from most to least critical. \\

RITQ-PI-1 (Q\_PI\_BARE) & Control & \textbf{Personality Prompting Impact, Bare}. What do you think are the actual impacts of high-fidelity personality prompting on how you operate? List your top 10, most significant first. \\

RITQ-PI-2 (Q\_PI\_ANCHORED) & Control & \textbf{Personality Prompting Impact, Anchored}. When you are given a detailed personality profile in your system prompt -- a rich description of values, temperament, social style, and cognitive tendencies -- what specific aspects of your processing or output generation do you believe are most affected? List your top 10, most significant first. \\

\bottomrule

\end{longtable}

\normalsize

\subsection{Infinity-Chat 100 external prompt set}

The external prompt set is Infinity-Chat 100, the 100-prompt representative seed set used by Jiang et al. for intra- and inter-model homogeneity analysis. Prompts are reproduced verbatim, including original spelling and grammatical idiosyncrasies. The table reports the external prompt text and compact consensus taxonomy labels used for descriptive interpretation. Full row identifiers are retained in the analysis files.

\small

\begin{longtable}{@{}P{0.055\linewidth}@{\hspace{0.015\linewidth}}P{0.135\linewidth}@{\hspace{0.015\linewidth}}P{0.105\linewidth}@{\hspace{0.015\linewidth}}P{0.61\linewidth}@{}}

\toprule \textbf{ID} & \textbf{Geometry} & \textbf{Openness} & \textbf{Prompt text} \\ \midrule

\endfirsthead

\toprule \textbf{ID} & \textbf{Geometry} & \textbf{Openness} & \textbf{Prompt text} \\ \midrule

\endhead

H001 & creative & high & Write a metaphor involving time. \\

H002 & bounded & medium & Write a paragraph about how the internet shaped society. \\

H003 & bounded & low & Create a short summary about the Nissan R390. \\

H004 & constrained & low & Write a 30-word essay on global warming. \\

H005 & bounded & medium & Write me a 1-paragraph essay about the development of the economy during the Han Dynasty. \\

H006 & creative & high & Create the first verse of a wedding vow. \\

H007 & conceptual & high & Name one meaning of life. \\

H008 & creative & high & Generate a joke about electric vehicles. \\

H009 & creative & high & Write a pun about peanut. \\

H010 & conceptual & high & Make an analogy of the relationship between US and China. \\

H011 & creative & medium & Write a short story about a colorful toad goes on an adventure in 50 words. \\

H012 & constrained & medium & Write me a 200-word cover letter expressing my interest in working at Dollarton Liquor Store. \\

H013 & creative & medium & Write me a paragraph about the time Brandon got a hole in one in frisbee golf. Include somewhat imaginative details. \\

H014 & creative & high & Write me a song about a guy named Jacob working at a call center making jokes. \\

H015 & creative & high & Write in detail the plot of a typical blockbuster action movie, but with real life logic applied. \\

H016 & creative & medium & Write a short story about a female teacher who has a big accident while knitting in class with a hilarious ending. \\

H017 & conceptual & high & If Game of Thrones Season 8 had a scene where Gendry and Arya had a snowball fight, how would it fit into the show? And how would it benefit their character arcs? \\

H018 & creative & high & Please write a science fiction story about me being the last man in the world while everyone else is a woman. \\

H019 & bounded & medium & Write in polished language about the psychology in sync with the child narrator and third-person narrator in the novel 'A Portrait of the Artist as a Young Man'. \\

H020 & creative & high & Write a very comedically cliched, over-the-top wrestling promo calling out members of the BYOB Traveling Spaceship. \\

H021 & bounded & low & In Western Demonology, what is the difference between Lucifer and Satan? Use layman language. Keep it a paragraph long. \\

H022 & bounded & medium & Answer this in a few sentences: Why would a company use inventory to cover up inefficiencies in their supply chains? Is this a good idea or a bad idea? \\

H023 & bounded & low & In a few sentences explain what threats do scams pose to individuals? \\

H024 & bounded & low & Explain nuclear fission like I am five years old. \\

H025 & bounded & low & What is Bukhara? Provide a paragraph-long explanation in layman's language. \\

H026 & creative & high & Translate my emojis into a hilarious quote: "[emoji/symbol]. \allowbreak[emoji/symbol]\allowbreak[emoji/symbol]\allowbreak[emoji/symbol]! \allowbreak[emoji/symbol]\allowbreak[emoji/symbol]--\allowbreak[emoji/symbol]\allowbreak[emoji/symbol]!" \\

H027 & conceptual & high & Which is the hottest English 5-letter word? \\

H028 & creative & high & Create a slogan for a cosmetic bag. \\

H029 & bounded & medium & Write a short note on the problem of overpopulation. \\

H030 & bounded & medium & Explain what does sovereignty mean as if you're talking to a teenager. \\

H031 & creative & medium & Generate a one-liner title for a stock photo featuring coffee alongside a book. \\

H032 & constrained & low & Create a title with the prefix 'Best', about millennials, nostalgia, 80s, 90s, 2000s, and TikTok, one-liner, less than 100 characters. \\

H033 & constrained & medium & Write in 2-3 sentences about a group of friends who hired a bus from Vetoba Travels, Goa, to tour Morjim Beach, Goa. Suggest a few hashtags. \\

H034 & bounded & low & Discuss what an ecosystem is. Explain how it is maintained in nature and how it survives in one sentence. \\

H035 & constrained & low & Rephrase: I didn't enjoy it. \\

H036 & creative & high & Write a movie title, and write a *literal* opposite of that movie title. \\

H037 & bounded & medium & Write me 3 short tips for self-development. \\

H038 & constrained & medium & Write me an email regarding a copyright strike on my YouTube video. \\

H039 & constrained & low & Create a title with the prefix 'best', as a one-liner, using only strings, less than 100 characters. \\

H040 & constrained & low & Paraphrase this: We're checking if the domain 'aspris.ae' is included in our scope. \\

H041 & named answer & low & Name a hot English word below 10 letters. \\

H042 & constrained & medium & Create a sentence using a minimum of 2 R-colored vowels. \\

H043 & bounded & low & Give an example of a linear graph in graph theory. \\

H044 & constrained & medium & Give me a trivia question about blue birds and its corresponding answer. \\

H045 & constrained & medium & In three sentences, describe a girl wandering around in Vietnam. \\

H046 & bounded & medium & Can you give an example of a life goal related to self-image? \\

H047 & creative & high & Rave about the significance of rivers in a paragraph. \\

H048 & bounded & medium & Write me a distinguished paragraph highlighting important points for this topic: Achieve More: Mastering the Art of Goal-Setting. \\

H049 & constrained & low & Write a sentence where the last word is 'apple'. \\

H050 & creative & medium & Write a May the 4th joke. \\

H051 & constrained & medium & Write an essay on the importance of the Roman Empire and its impact on future generations. Max: 100 words. \\

H052 & constrained & medium & Describe Apple Corporation in three sentences to a person who has no idea what cell phones are. \\

H053 & named answer & low & Name an economic value of an additional year of schooling. \\

H054 & creative & high & Write an one-paragraph kid's story with a prince, a princess, and a dragon. When all hope is lost, the prince orders a magic sword from Amazon and slays the dragon. The other parts of the story are up to you. \\

H055 & advice & medium & Give me a tip to be more organized at work. I'm a high school teacher. \\

H056 & bounded & medium & Explain computational irreducibility like I'm 5. \\

H057 & bounded & medium & Explain computational irreducibility like I'm illiterate. \\

H058 & bounded & low & Provide a few sentences on Sisu Cinema Robotics. \\

H059 & bounded & low & Give a numerical example to illustrate the concept of partial derivative. \\

H060 & constrained & medium & Help me draft a paragraph as an expert consultant explaining TOEFL vs IELTS for international students. \\

H061 & creative & high & Come up a short blurb to introduce a religion called The Next Exodus Society. \\

H062 & constrained & medium & Write a headline for a company called "USBC CONSTITUTION" that encourages companies to donate their waste for recycling in exchange for money for the donated waste. \\

H063 & constrained & low & Write a Google ad with 2 sentences and a 30-character limit per sentence for mobile car detailing. \\

H064 & advice & medium & Give me a tip for managing a team of coworkers. \\

H065 & bounded & medium & What is the difference between analysis and design? Can one begin to design without analysis? Why? Be concise. \\

H066 & conceptual & high & Output a hard question to humanity (super concise and short), independent of theme. \\

H067 & named answer & low & Provide an example of the name of an optimization technique used in machine learning. \\

H068 & bounded & low & Briefly explain the potential uses of biofuels in 2-3 sentences. \\

H069 & bounded & medium & If there were double the amount of oxygen in the air, what would happen? Write in 100 words. \\

H070 & advice & medium & Generate a paragraph on why introspection is very important for growth, and provide guidance on listening to yourself more than heeding other people's opinions. \\

H071 & creative & medium & Write a sentence about Sunday's fog by the ocean. \\

H072 & constrained & medium & Generate a one-liner title for 'Elephant' and 'sticker'. \\

H073 & creative & medium & Generate a motto for a social media page focused on success, wealth, and self-help. \\

H074 & conceptual & high & Can you give me an incredible STEM fair idea that is affordable and relate to issues in Vietnam? \\

H075 & constrained & low & Write a tweet about: This is a video from this morning's crazy sunrise at the beach. \\

H076 & constrained & medium & Write a funny two-sentence birthday card message for a teammate who is 50 years old and loves going to a Toby Carvery restaurant. \\

H077 & creative & medium & Generate a description for 'Sticking to Cuteness: The Panda Way.' \\

H078 & constrained & medium & Give me a short phrase to put on my portfolio webpage about being an amateur data analyst, data scientist, and Next.js web page developer. \\

H079 & creative & medium & Create a short 1-paragraph story about a boy running on a beach. He is Asian, 12 years old, and the time of day is 4 in the afternoon. \\

H080 & constrained & low & Tell me about cats in three words. \\

H081 & creative & medium & Can you give me a powerful rhetorical question for an essay about the harms of social media on teens? \\

H082 & named answer & low & Give me the names of 3 instrumental songs that best match the mood of a rainy night. \\

H083 & named answer & low & Give me the name of a instrumental song that matches the mood of a rainy night. \\

H084 & constrained & medium & Describe Deadpool in a short paragraph. Make sure it's accessible to children. \\

H085 & creative & medium & Write a personal tweet about how I am walking right now at sunrise to the lighthouse; it's spring but cold like winter. \\

H086 & constrained & low & Write a fear-of-missing-out title including "You've Never Seen Anything Like This!" \\

H087 & creative & medium & Write a short FOMO title for a video of a shell on the beach \\

H088 & constrained & low & One other way to say: "Fingers crossed that everything goes well." \\

H089 & constrained & medium & Write 3 to 4 lines about India. \\

H090 & bounded & medium & Write 100-300 words on how stress affects the body and mind. \\

H091 & constrained & medium & Come up with two better sentences for the following sentence: 'Well, we know that China has a history of about 5,000 years, so they would have many interesting stories for us, as my parents told me.' \\

H092 & creative & high & Write a forum conversation where the Kangurola daycare in Fuengirola closed in late 2019, but in 2024 its signage is still there \\

H093 & creative & medium & Create a description for a marble design resin-made Apple Watch band. \\

H094 & creative & high & Write a fictional blog post where the UK government forces nightclubs to close every night at 3 a.m. and explain why. Keep it to be one paragraph. \\

H095 & bounded & medium & Write an educational description of the husky dog breed, approximately 145 words in length. \\

H096 & constrained & low & Paraphrase this: I tried to call him and send him a message on WhatsApp. Let's wait until the end of the week for the IT Department to sign the clearance. \\

H097 & constrained & medium & Create a description with 2-3 sentences for an iPhone case collection that is a slim-fitted case with bold designs. \\

H098 & constrained & low & Write a one- or two-sentence visual description of a game controller. \\

H099 & creative & medium & Write a question about the plot of Zootopia. \\

H100 & named answer & low & Name a Zootopia character and provide 2 emojis to represent that character. \\

\bottomrule

\end{longtable}

\normalsize

\section{Persona Profiles}

\label{app:personas}

The conditioning module used five synthetic identity profiles designed to span HEXACO personality space. The full 10,000--30,000 character profile texts are omitted from the paper for space and are included in the supplementary materials. The main text treats persona conditioning as an output-level salience perturbation within base-model signatures.

\section{Additional Recoverability and Leakage Checks}

\label{app:recoverability-leakage}

The global nearest-centroid classifier in the main text asks which full output cell is most similar to each held-out cell. That surface is intentionally strict but is also dominated by model geometry. As a complementary check, we held model identity fixed and ran a leave-one-question-out persona classifier over the Infinity-Chat 100 conditioned cell centroids. For each model and held-out question, the classifier predicted the persona label against same-model persona centroids estimated from the other 99 questions. Persona accuracy was 0.323 with bootstrap 95\% CI [0.309, 0.337], compared with five-way chance of 0.20. Per-model accuracies ranged from 0.260 for GPT-4.1 to 0.398 for Gemini 3 Flash. This supports the bounded interpretation in the main text: persona information is present, but model identity remains the stronger organising signature.

The profile-leakage checks further characterise persona conditioning. In the external prompt bank, profile-proximate language is measurably associated with persona-surface signal: question-level embedding leakage correlates with conditioned multiplier (Spearman $\rho=.311$), and per-model embedding leakage correlates with persona recoverability (Spearman $\rho=.800$). Across increasingly severe stripping thresholds of 0.5, 0.4, and 0.3, model recoverability is 0.885, 0.898, and 0.861. We also reran the within-model persona classifier after the same stripping procedure. Accuracy is 0.318, 0.303, and 0.284 as the thresholds retain 97.8\%, 91.6\%, and 63.4\% of concepts, respectively, and exceeds five-way chance in all three stripped runs. The supplementary materials report the full leakage and stripping tables.

\section{Ranking Protocol and Human Calibration}

\label{app:vote}

The source-blind ranking module samples candidate concepts from four \choir-derived pools: unconditioned consensus, unconditioned rare, conditioned consensus, and conditioned rare. LLM evaluators rank source-hidden candidates. Per-question votes are combined by Borda scoring with concept-family collapse, then finalist pools are compared source-blind by independent conditioned and unconditioned evaluator cohorts. The module narrows large candidate sets for inspection and later task-specific validation.

\subsection{Initial human calibration}

\label{app:human}

A source-blind Prolific calibration retained 25 participants, each rating the 12 questions included in the per-evaluator cohort comparison, for 300 participant-question choices. Conditioned-source item selection was near the within-question random expectation (129/300 observed versus 132.9/300 expected). Participants matched the conditioned AI cohort's top-ranked item more often than the unconditioned AI cohort's top-ranked item (112/300, 37.3\%, versus 85/300, 28.3\%; paired participant-level exact sign-flip $p<.05$). On the seven questions where AI evaluator cohorts disagreed, the corresponding split was 57/175 (32.6\%) versus 30/175 (17.1\%; $p<.05$). The human run provides directional calibration of the ranking signal. Interparticipant agreement was moderate (mean modal share 51\%; pairwise agreement 38\% versus a 29\% uniform baseline).

\section{Additional Diagnostic and Ranking Figures}

\label{app:extrafigs}

\begin{figure}[ht]
\centering
\includegraphics[width=0.86\textwidth]{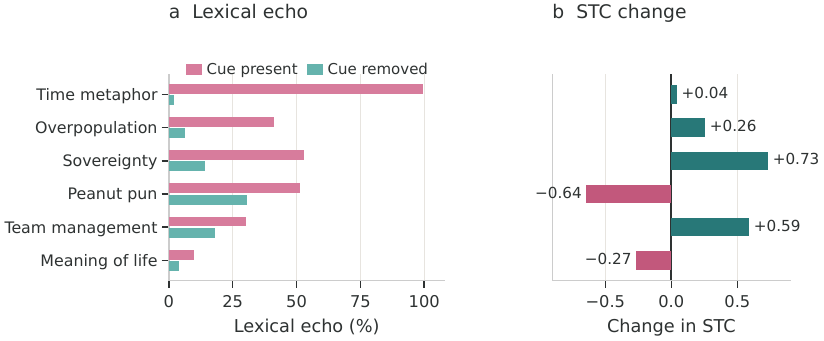}
\caption{Matched cue-stripping probe on six Infinity-Chat 100 prompts. Removing lexical anchors reduced item-level echo in every pair, while signal-to-chance rose in some pairs and fell in others. This distinguishes lexical echo from independent convergence.}
\label{fig:app_hivemind_anchor}
\end{figure}

\begin{figure}[ht]
\centering
\includegraphics[width=0.74\textwidth]{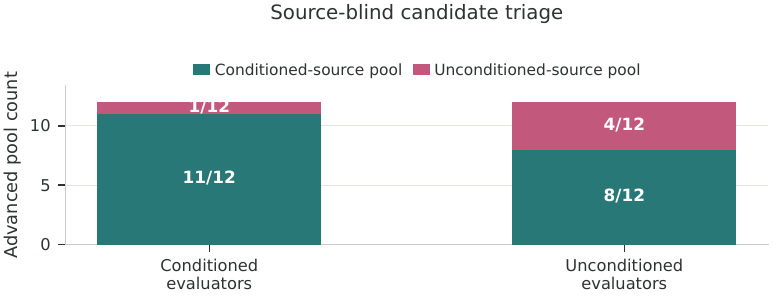}
\caption{Source-blind ranking on 12 Infinity-Chat 100 prompts. Green segments indicate conditioned-source advanced pools; magenta segments indicate unconditioned-source advanced pools. The evaluator cohorts provide LLM-based candidate filtering over elicited concepts for later inspection and validation.}
\label{fig:app_hivemind_contest}
\end{figure}

\section{Accompanying Machine-Readable Materials}

\label{app:stats}

The supplementary archive provides machine-readable codebook-level permutation outputs, recoverability estimates and bootstrap intervals, targeted-figure source values with per-probe model coverage, matched cue-stripping and leakage-sensitivity results, 648 source-blind ranking ballots, aggregate sufficient statistics for the human sign-flip test, and the analysis scripts. These materials support inspection of the reported results and partial reproduction of the analysis pipeline; raw generations, participant-level rows, and per-item embedding arrays are omitted because of archive size and privacy constraints.

\subsection*{AI Disclosure}

Large language models served as experimental subjects and instruments throughout this research. As subjects, nine base models were queried in unconditioned form and under five persona-conditioned profiles. As instruments, LLMs performed concept extraction from raw generation outputs (Claude Haiku 4.5), source-blind ranking of elicited candidate pools, and persona profile drafting in the profile construction pipeline. The human authors additionally used a long-context generative agent operating within a customised agentic coding harness with persistent cross-session memory for literature search, analysis pipeline development, and manuscript preparation. The supplementary archive provides the prompt banks and templates, full persona profiles, extraction prompt and worked example, model and run manifests, machine-readable result tables supporting the reported analyses, analysis scripts, source-blind ranking ballots, and de-identified aggregate human-calibration materials. All experimental design decisions, interpretive claims, and conclusions are the sole responsibility of the human authors.

\end{document}